%% file: paper.tex
\documentclass[conference]{IEEEtran}

\usepackage{microtype}
\usepackage{graphicx}
\usepackage[caption=false,font=scriptsize,labelfont=sf,textfont=sf]{subfig}
\usepackage{booktabs}
\usepackage{cite}
\usepackage{amsmath,amssymb}
\usepackage{amsthm}
\usepackage{xcolor}
\usepackage{soul}
\usepackage[normalem]{ulem}
\usepackage{makecell}
\usepackage{xurl}
\usepackage[hidelinks]{hyperref}

\newtheorem{definition}{Definition}
\newtheorem{theorem}{Theorem}

\newtheorem{corollary}{Corollary}

\title{Robust Privacy: Inference-Stage Privacy through Certified Robustness}

\author{
\IEEEauthorblockN{
Jiankai Jin,
Xiangzheng Zhang,
Zhao Liu,
Wenzhuo Xu,
Dongdong Yang,
Deyue Zhang,
and Quanchen Zou
}
\IEEEauthorblockA{
360 AI Security Lab\\
\{jinjiankai, zhangxiangzheng, liuzhao3, xuwenzhuo, yangdongdong1, zhangdeyue, zouquanchen\}@360.cn
}
}

\begin{document}

\maketitle

\begin{abstract}
An adversary observing a model's released prediction can infer sensitive
attributes of the queried input, or even reconstruct representatives of the
model's training data. The inference interface thus acts as a side channel for
privacy leakage. We introduce Robust Privacy (RP), an inference-stage privacy
notion inspired by certified robustness: if a model's prediction is provably
invariant within a radius-$R$ neighborhood around an input $x$ with confidence
at least $1-\alpha$, then $x$ enjoys $(R,\alpha)$-Robust Privacy, under which
we prove that any adversary observing the released prediction has at most
$\alpha/2$ advantage in distinguishing $x$ from any input within distance 
$R$ of~$x$. Building on RP, we formalize Robust Attribute Privacy (RAP), an 
attribute-level privacy notion that characterizes the set of sensitive-attribute 
values that remain compatible with a released prediction when the adversary's 
side information is fixed. On a classification task with a sensitive attribute, 
RP increases the median length of the RAP-compatible inference interval from 
$23.50$ to $29.96$, reducing attribute-inference precision. Model inversion 
attacks, often treated as a training-stage threat, in fact rely on fine-grained 
input--output dependence signals leaked through the inference interface; RP 
masks these signals at the inference stage, reducing attack success rate (ASR) 
from $73\%$ to $4\%$ on a black-box inversion attack. This direct targeting of 
the leakage channel enables RP to dominate DP-SGD and randomized response in 
the privacy--utility tradeoff space: RP retains $98.4\%$ accuracy at $21\%$ ASR, 
whereas DP-SGD must drop accuracy to $61.7\%$ to reach a comparable ASR. Across 
both experiments, increasing the smoothing sample size $N$ at fixed noise scale 
strengthens privacy and improves utility together, at higher per-query inference 
cost. Finally, we examine model distillation as a scope boundary and show that 
RP mitigates attribute-level and instance-level inference-stage privacy leakage, 
but not function-level extraction through~model~distillation.
\end{abstract}

\input{intro}

\input{related}

\input{threat}

\input{privacy}

\input{exp_attribute}

\input{exp_instance}

\input{exp_distillation}

\input{discuss}

\section{Conclusion}

The locus of privacy enforcement in machine learning has, by convention, been
the training stage. Differential privacy and its descendants control what a
learned model can memorize about any individual record, based on the premise
that if memorization is bounded at training time, downstream leakage will also
be bounded. This paper argues that this premise leaves a real attack surface
uncovered. Attribute inference, model inversion, and related attacks do not
succeed only because the model has memorized too much; they also succeed because
the inference interface itself is informative: released predictions can reveal
fine-grained information about queried inputs. A defense that only constrains 
training cannot, by construction, directly suppress this privacy leakage side 
channel through the inference interface.

Robust Privacy is a concrete step toward treating the inference interface as a
privacy boundary in its own right. Its mechanism is simple: certified prediction
invariance in a neighborhood of the queried input is reinterpreted as an
inference-stage indistinguishability guarantee under the released prediction.
It shows that inference-stage privacy does not require a new cryptographic 
primitive or a new training pipeline, but can instead arise from reinterpreting 
a guarantee that the robustness community has already spent a decade building. 
Robust Attribute Privacy extends this reinterpretation to the attribute level: 
a wider set of sensitive values is now compatible with the same released label.

We see two directions in which this view becomes productive. First, the notion
of an indistinguishable neighborhood under a released prediction is not tied to
randomized smoothing; any certification method (e.g., deterministic verifiers,
smoothing-based wrappers, or future techniques for large generative models) can
instantiate RP, and each will trade off privacy, utility, and inference cost
differently. Second, RP and training-stage defenses are complementary rather
than competing: a model trained under DP and queried under RP would enjoy
bounded memorization and bounded inference interface leakage simultaneously.
Privacy in deployed AI systems will increasingly be decided at the inference
boundary, where the user actually meets the model. Robust Privacy is one notion
of what it means to defend~that~boundary.

\section*{Acknowledgment}

Generative AI was used for editorial purposes in this manuscript, including
language polishing, grammar checking, and improving clarity of presentation.

\section*{Ethics Considerations}

This work studies privacy leakage and defenses at the inference stage of
machine learning systems, and the method we propose, Robust Privacy, is a
privacy-enhancing defense. All experiments are offline and use public
benchmark datasets and locally trained models; we do not interact with live
systems, deployed services, or users, and we collect no new human-subject
data, so institutional ethics review was not applicable. Although the
experiments involve human-related attributes and public face-image
benchmarks, we report only aggregate metrics and make no attempt to identify
individuals, link records to real persons, or release per-person sensitive
inferences. The attacks we evaluate (attribute inference, model inversion,
and model distillation) are existing techniques from prior published work,
used solely to measure leakage and the effectiveness of the defense; we
introduce no new attack capability, and our results expose no vulnerability
in any deployed system, so no responsible disclosure was required.

\bibliographystyle{IEEEtran}
\bibliography{refs}

\end{document}

%% file: intro.tex
\begin{figure}[!t]
    \centering
    \subfloat[
        Attribute inference, an attribute-level attack. A health service answers 
        ``Insulin Recommended'' for a patient record. The four silhouettes depict the
        same record with only the BMI value changed, and the star marks the true one.
        Without RP, the answer can pin the sensitive attribute near 35. With RP,
        a certified range of sensitive values receives the
        same ``Insulin Recommended'' answer, so every candidate in that range remains
        plausible, and hence indistinguishable, under the released prediction.
    ]{%
        \includegraphics[width=\linewidth]{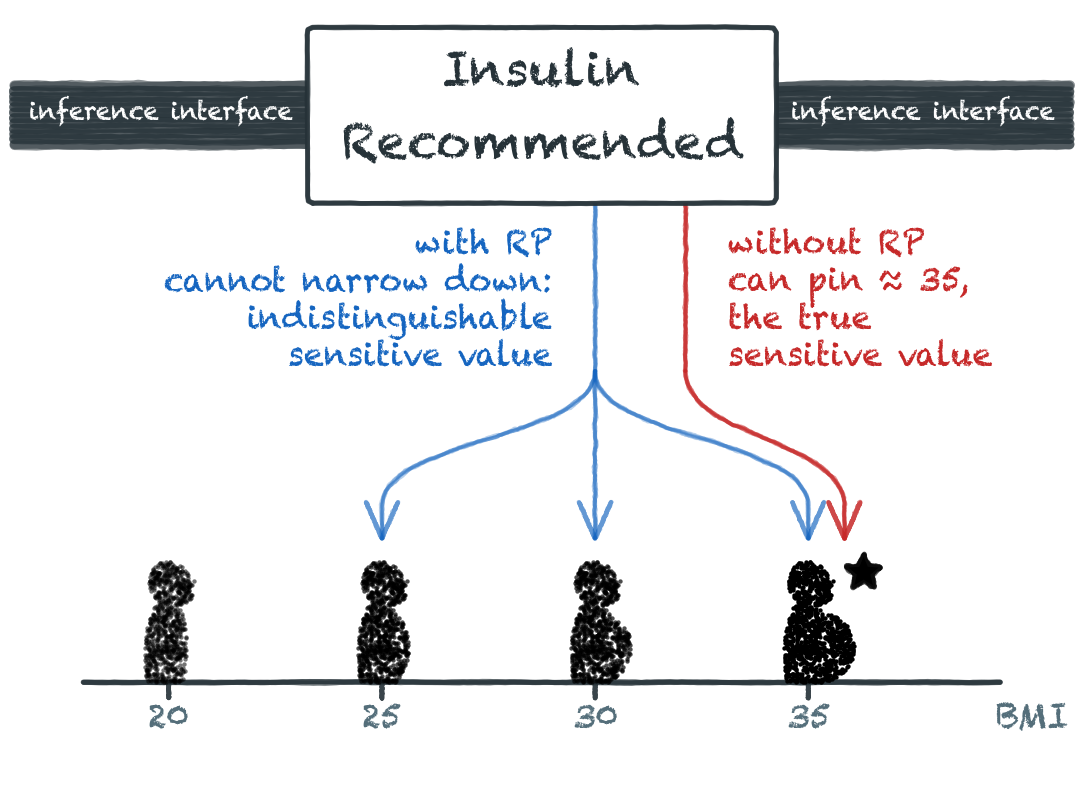}%
        \label{fig:teaser-attr}}
    \\
    \subfloat[
        Model inversion, an instance-level attack. An attacker queries the model over
        many rounds and refines a synthetic input using the released prediction (e.g.,
        whether ``cat'' or not). Without RP, a representative image of the target
        class in the training distribution can be reconstructed; with RP, the
        reconstruction fails.
    ]{%
        \includegraphics[width=\linewidth]{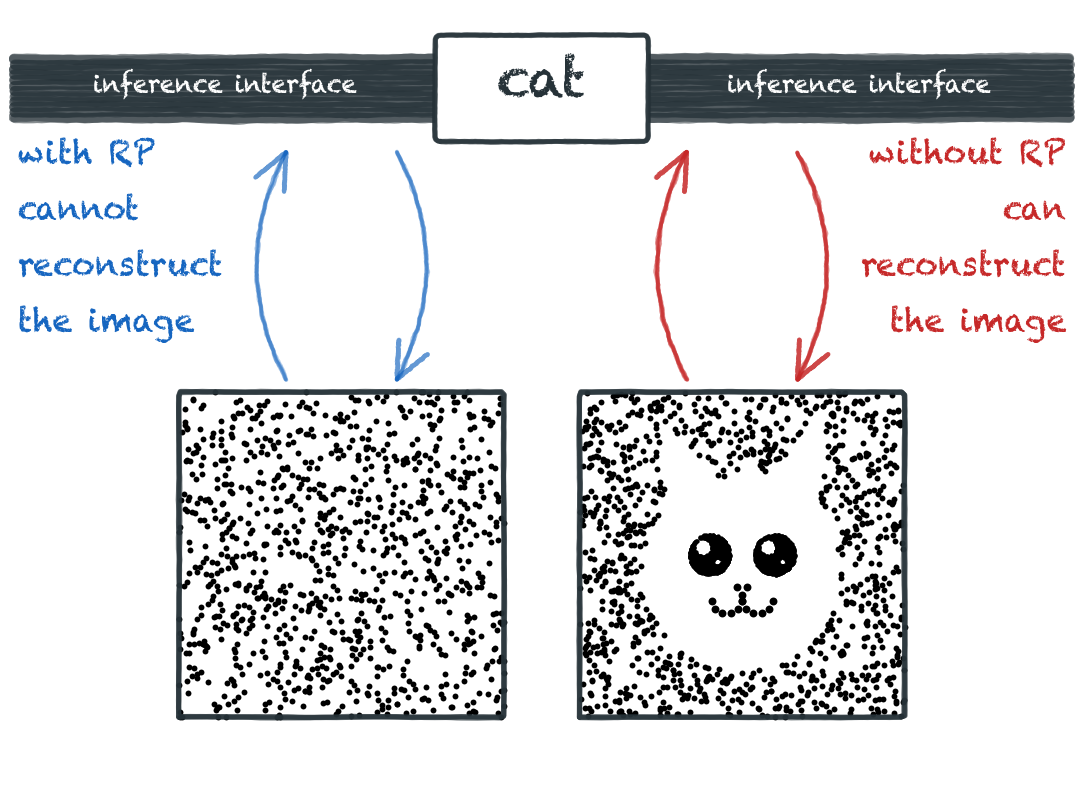}%
        \label{fig:teaser-inv}}
    \caption{
        Robust Privacy (RP) mitigates privacy leakage
        through the inference interface, as illustrated on the
        attribute-level and instance-level attacks evaluated in this paper.
    }
    \label{fig:teaser}
\end{figure}

\section{Introduction}
\label{sec:intro}

Consider a user querying a personalized health-management service on a mobile
device. The service returns a recommendation (e.g., whether insulin therapy is
suggested for this user). The recommendation is a prediction released through
the inference interface of the personalized model, and can be captured via
screen sharing or telemetry logs accessible to a third
party~\cite{pan2018panoptispy,ftc2023goodrx}. An adversary with
this prediction sees nothing of the model's parameters, training data, or even
its confidence scores, but may obtain side information about the user (e.g., from
a public online profile): approximate age range, gender, height, and other profile
attributes. Given this side information, the observed recommendation constrains
the remaining clinically relevant value (e.g., body mass index) that could have
produced it, especially when the model's decision depends strongly
on that value after conditioning on the known attributes. From the released
recommendation and the available side information, the adversary can therefore
narrow to a tight interval the remaining sensitive attribute, which the
user never disclosed and the service never released
(Figure~\ref{fig:teaser-attr}).

The same inference interface also supports a second kind of adversary. This
adversary submits synthetic queries and observes how the released prediction
changes in response, using that feedback to iteratively refine the queries
toward the training distribution and eventually recover representatives of
the records the model was trained on (Figure~\ref{fig:teaser-inv}). In both
cases, the attacker's shared channel is the model's inference interface: the
prediction returned by the model.

This kind of side-channel leakage through the inference interface is not fully
addressed by defenses currently in deployment. Differential privacy, typically 
applied during training~\cite{abadi2016deep}, bounds how much any individual 
training record influences the learned model, but it does not directly bound
what each released prediction at query time reveals about the queried input,
or about the training distribution as probed by an attacker's synthetic
queries. The inference interface is, by construction, informative: the prediction 
has to depend on the input, or the service is useless. That dependence is exactly 
what attribute inference~\cite{fredrikson2014privacy,zhao2021feasibility,
mehnaz2022your,jayaraman2022attribute,mireshghallah2023can} and model inversion
attacks~\cite{fredrikson2015modelinversion,yang2019adversarial,aivodji2019gamin,
zhang2020secret,chen2021knowledge,kahla2022label} exploit. We argue that the
inference interface deserves to be treated as a first-class privacy boundary,
with its own notion of what it means for a released prediction to leak too much,
and its own mechanisms for enforcement. This paper proposes one such notion:
Robust Privacy.

Robust Privacy (RP) draws on a guarantee that the robustness community has 
spent the last decade formalizing. Certified robustness~\cite{lecuyer2019certified,
li2019certified,cohen2019certified} certifies that a classifier's prediction
remains unchanged for all inputs within a neighborhood of a queried input,
with the neighborhood defined by a robust radius $R$ under the $\ell_p$ norm
(e.g., $p\in\{1,2,\infty\}$). Techniques such as randomized
smoothing~\cite{cohen2019certified} make this guarantee practical for modern
deep networks. We repurpose this guarantee for inference-stage privacy: if an
adversary cannot distinguish an input $x$ from a nearby input $x'$ based on
the model's identical prediction, then releasing that prediction does not leak
fine-grained information about $x$ that distinguishes it from nearby inputs
during inference. We formalize this invariance-based notion of inference-stage
privacy as Robust Privacy. The resulting notion admits a standard indistinguishability 
semantics: any adversary observing the released prediction has provably bounded 
advantage in distinguishing the queried input from any other input within its 
certified $R$-neighborhood.

The rest of the paper develops this notion along three axes. Conceptually,
Section~\ref{sec:rp-def} formalizes RP using the robust radius $R$ as its
privacy metric: a larger $R$ corresponds to a larger neighborhood around the
queried input in which nearby inputs are indistinguishable under the released
prediction. The section then formalizes Robust Attribute Privacy (RAP) as the
corresponding attribute-level notion, characterizing the set of
sensitive-attribute values compatible with the released prediction when 
the adversary's side information $x_{-1}$ is fixed. Operationally,
Section~\ref{sec:threat} defines three inference-interface attack
settings: attribute inference, model inversion, and model distillation.
The first two are the primary privacy-leakage settings that RP is designed
to mitigate, while the third is used as a boundary case. Empirically,
Sections~\ref{sec:moti} and~\ref{sec:instance-exp} show that RP delivers
concrete protection in the first two settings: a reduction in 
attribute-inference precision and a drop in the success rate of a black-box 
inversion attack. Section~\ref{sec:distillation} then examines distillation 
as a boundary case, delineating what RP, as an inference-stage privacy notion 
based on certified invariance, can and cannot protect.

We do not position RP as a replacement for training-stage privacy mechanisms.
RP and Differential Privacy (DP)~\cite{dwork2006calibrating}, often
instantiated in machine learning via differentially private stochastic gradient
descent (DP-SGD)~\cite{abadi2016deep}, operate at different stages of the
model's life cycle, target different information channels, and use different
mechanisms. DP-SGD bounds what the trained model memorizes about any single
training record, and it remains the right defense for that channel. The
inference interface exposes a separate channel: each released prediction
discloses information about the queried input (e.g., which attribute value 
produced it) and this disclosure is not what DP-SGD is designed to bound. 
In this paper, we show that, for this channel, an inference-stage defense 
can suppress leakage that a training-stage defense cannot reach at comparable 
utility. We return to the comparison and composition of RP and DP in the discussion.

Our contributions are:
\begin{itemize}
\item \textbf{Robust Privacy.}
We introduce Robust Privacy (RP), an inference-stage privacy notion that
reinterprets certified robustness as an indistinguishability guarantee
under the released prediction: when a model's prediction at $x$ is
certifiably invariant within a radius-$R$ neighborhood with confidence
at least $1-\alpha$, we say $x$ enjoys $(R,\alpha)$-Robust Privacy, and
we prove that any adversary observing the released prediction has
distinguishing advantage at most $\alpha/2$ between $x$ and any neighbor
$x'$ within distance $R$. RP thus shifts the privacy boundary to the
inference interface, with a quantitative bound on what each released
prediction can leak about its input through this side channel.

\item \textbf{Robust Attribute Privacy.}
We formalize Robust Attribute Privacy (RAP), which characterizes the set
of sensitive-attribute values that remain compatible with a released
prediction when the adversary's side information $x_{-1}$ is fixed.
We further prove a direct implication from RP to RAP: whenever $x$
satisfies $(R,\alpha)$-RP, the RAP-compatible set provably contains a
sub-interval of length $2R$ around the true sensitive attribute value 
(with confidence $1-\alpha$), giving a certified lower bound on the
attribute-level uncertainty the attacker faces.

\item \textbf{Attribute-level evaluation: RP mitigates attribute inference.}
On a classification task with a continuous sensitive attribute, we instantiate
RP via randomized smoothing and empirically validate the RAP-compatible
interval expansion under an attribute-inference attack setting: as the noise
scale increases, the median length of the compatible sensitive-attribute
inference interval increases from $23.50$ under the unprotected baseline to
$29.96$ under RP, reducing attribute-inference precision. Increasing the
smoothing sample size $N$ at fixed noise scale widens this interval and raises
test accuracy together, indicating that RP can strengthen privacy and improve
utility simultaneously at higher per-query inference~cost.

\item \textbf{Instance-level evaluation: RP mitigates model inversion.}
On a black-box model inversion attack against a face classifier, RP empirically
reduces attack success rate (ASR) from $73\%$ to $4\%$. RP dominates DP-SGD and 
randomized response in the privacy--utility tradeoff space on the same inversion 
attack. RP retains $98.4\%$ accuracy at $21\%$ ASR, whereas DP-SGD must drop accuracy 
to $61.7\%$ to reach a comparable ASR. The gap is structural: DP-SGD constrains 
training-stage memorization, while the attack signal is exposed through the 
inference interface; an inference-stage defense is therefore better positioned 
to suppress that signal.

\end{itemize}

%% file: related.tex
\section{Related Work}
\label{sec:related}

This work brings together research lines that are usually studied separately:
attacks that exploit the inference interface, and privacy mechanisms
that defend against training-stage or inference-stage leakage. We 
review each line below and discuss its connection with Robust Privacy.

\subsection{Attribute Inference Attacks}

Fredrikson et al.~\cite{fredrikson2014privacy} introduced the concept of
attribute inference: given a pharmacogenetic model for warfarin dosing, an
adversary uses the model output (i.e., the predicted dosage) together with
known background attributes to infer sensitive genetic markers. Their results
show that model outputs can leak sensitive input attributes during inference.
Subsequent work studied attribute inference in more general ML settings,
including systematic analyses of when such inference is feasible and when it
reduces to standard imputation~\cite{zhao2021feasibility,mehnaz2022your,
jayaraman2022attribute}. These studies typically assume that all attributes
other than the target sensitive attribute are known, consistent with Scenario~I
in Section~\ref{sec:threat}. Many attacks exploit richer soft model outputs
(e.g., confidence vectors or regression values), while some works consider
label-only access~\cite{mehnaz2022your}. In Section~\ref{sec:moti}, we design
a label-only, black-box attribute inference attack and evaluate Robust Privacy
as an inference-stage privacy mechanism for reducing attribute-inference
precision (i.e., expanding the RAP-compatible sensitive-attribute inference~interval).

While this line of work is sometimes viewed as an instance of model inversion,
we treat it as attribute inference because it recovers only a subset of a query
subject's attributes. In contrast, we use model inversion to refer to attacks
that optimize over the input space to reconstruct representative samples from 
the model's training distribution.

\subsection{Model Inversion Attacks}
\label{sec:related:mia}

Model inversion attacks aim to reconstruct representative inputs that 
reveal information about a model's private training data. Fredrikson et
al.~\cite{fredrikson2015modelinversion} formalized model inversion using
confidence values and showed that recognizable face images can be recovered
with black-box access. Subsequent black-box attacks that exploit soft outputs
improve inversion quality by aligning auxiliary knowledge~\cite{yang2019adversarial}, 
jointly training a surrogate and a generator~\cite{aivodji2019gamin}, 
optimizing in a GAN latent space with a public-data prior~\cite{zhang2020secret}, 
or distilling target-model knowledge into the generator~\cite{chen2021knowledge}. 
Kahla et al.~\cite{kahla2022label} further showed that inversion remains 
possible in the stricter label-only black-box setting, by probing small 
perturbations and tracking label changes to estimate an optimization signal.

A key enabler of these inversion attacks is that model outputs during
inference remain sensitive to small input perturbations, providing an
iterative optimization signal through confidence or label changes exposed 
by the inference interface. Robust Privacy is designed to mask this 
fine-grained inference-stage signal by establishing an indistinguishable 
neighborhood around the queried input under the released prediction.

\subsection{Model Distillation}
\label{sec:related:distillation}

Model distillation was originally introduced to transfer knowledge from a
cumbersome teacher model, often an ensemble or a large network, to a smaller
student~\cite{hinton2015distilling}, but it has also been adapted as a model
extraction primitive: by querying a deployed model and training a student on
its predictions, an adversary can approximate the deployed model without access
to its parameters~\cite{tramer2016stealing,papernot2017practical}. Subsequent
attacks extend this idea to deployed deep networks, as well as to data-free
settings where the adversary synthesizes queries instead of relying on natural
in-distribution data~\cite{orekondy2019knockoff,jagielski2020high,kariyappa2021maze,
truong2021datafree}. DisGUIDE~\cite{rosenthal2023disguide}, which we evaluate 
in Section~\ref{sec:distillation}, further performs hard-label data-free 
distillation by using disagreement between students to guide query generation.

We evaluate model distillation as a scope boundary for RP rather than as a
primary target: per-query input--output masking does not directly counter
aggregate function-level extraction. We elaborate on this scope distinction
in Sections~\ref{sec:threat} and~\ref{sec:discus}.

\subsection{Privacy-Preserving Machine Learning}
\label{sec:related:ppml}

Differential Privacy (DP)~\cite{dwork2006calibrating} is a privacy framework
for limiting the influence of individual records on released computations.
In machine learning, a common instantiation is DP-SGD~\cite{abadi2016deep},
which clips per-example gradients and adds noise during training to reduce 
the amount of training-record information memorized by the learned model, 
thereby limiting privacy leakage through the resulting model.

Randomized response~\cite{warner1965randomized,opendp_randomized_response} is
another DP mechanism, originally designed for collecting sensitive categorical
responses. Each categorical value is randomized before release; for a hard-label
prediction API, the released label can be treated as such a categorical response.
This gives a simple output randomization method at inference: after the base model 
predicts a label, the API releases the predicted label with some probability and 
otherwise releases a random alternative label. One adaptation of randomized
response to privacy-preserving machine learning is BDPL~\cite{zheng2019bdpl},
which uses boundary randomized response to obfuscate predictions near the
decision boundary against model distillation attacks.

RP differs from both mechanisms. Unlike DP-SGD, RP is applied at the 
inference stage and does not require retraining the target model. Unlike 
randomized response, RP does not randomly flip each released label; 
instead, it uses randomized smoothing to produce labels that are stable 
within an indistinguishable neighborhood of the queried input. This 
masks fine-grained input--output dependence signals leaked through the 
inference interface, and helps stabilize model utility by aggregating 
predictions over Gaussian-perturbed copies of the queried input.

\subsection{Certified Robustness}
\label{sec:related:cr}

Certified robustness studies whether a classifier's prediction is provably
invariant within a neighborhood of a given input. Concretely, a robustness
certificate for an input $x$ provides a robust radius $R$ such that the
prediction remains unchanged for all inputs $x'$ with $\|x'-x\|_p \le R$ 
(e.g., $p \in \{1,2,\infty\}$). In the robustness literature, this rules out 
the existence of adversarial examples~\cite{42503,goodfellow2014explaining,
carlini2017towards,madry2017towards} within that neighborhood. Existing 
certification approaches fall into two categories: deterministic methods 
that conservatively over-approximate network behavior (e.g., bound
propagation~\cite{crown,beta-crown}, convex relaxations, and mixed-integer
formulations~\cite{ehlers2017formal,tjeng2017evaluating,bunel2020branch}), 
and probabilistic methods via randomized smoothing~\cite{cohen2019certified,
li2019certified}, which certify a robust radius up to a user-chosen failure 
probability~$\alpha$.

In this work, we instantiate Robust Privacy using the smoothing framework of
Cohen et al.~\cite{cohen2019certified}, as it is a widely used certification
method with standard $\ell_2$ certificates and enables direct control of the
inference-stage privacy level by tuning its parameters (e.g., noise scale
$\sigma$ and Monte Carlo sample size $N$).

\subsection{Privacy and Robustness}
\label{sec:related:privacy}

Prior work has connected privacy and robustness, though the connection differs 
from the one made by RP. PixelDP~\cite{lecuyer2019certified} injects noise to 
obtain DP-style prediction stability and derive adversarial robustness 
certificates. DP-CERT~\cite{wu2024dpcert} combines DP training with robustness 
certification by integrating augmentation-based smoothing into DP-SGD. 
Shredder~\cite{mireshghallah2020shredder} protects inference-stage privacy in 
split edge--cloud inference by learning additive noise distributions for 
transmitted intermediate representations, reducing their information content 
while preserving inference accuracy.

A separate theoretical line connects DP with statistical robustness at the
population-estimation level~\cite{dwork2009differential,hopkins2023robustness,
asi2023robustness}, attaching privacy to the estimator rather than to each
released~prediction.

RP differs from these works in two ways. First, it operates through the 
inference interface rather than at the learning stage: DP-CERT and the 
algorithmic-statistics line all attach privacy to the model or the learning 
algorithm before deployment, whereas RP's guarantee is attached to each 
individual query at inference time. Second, RP repurposes the direction 
of the robustness--privacy connection: PixelDP uses differential privacy 
to certify robustness, and DP-CERT combines DP training with robustness 
certification, while RP uses certified robustness to provide privacy for 
that queried input. Shredder shares the inference-stage motivation but 
protects intermediate representations exposed by split deployments, a 
different surface from the inference interface RP addresses.

%% file: threat.tex
\section{Threat Model}
\label{sec:threat}

We consider a black-box adversary who can observe model predictions
(hard labels) but has no access to confidence scores, gradients, or internal
model parameters. The adversary interacts with the model through the inference
interface in three ways: (i) submitting inputs with varying candidate values
for a missing sensitive attribute while holding known attributes fixed, and
observing the released predictions to infer the missing attribute value; 
(ii) submitting synthetic inputs and using prediction feedback to iteratively
optimize them toward the training data distribution; and (iii) querying many
inputs to learn a global approximation of the model's input--output function.

\paragraph{Scenario I: Sensitive Attribute Inference}
In this setting, the attacker targets a specific user during inference and
seeks to infer a sensitive attribute $x_1$ (e.g., BMI or age) from the model's
released prediction.

To formalize side information, we let $x_{-1}$ denote all attributes other 
than the sensitive target attribute $x_1$, and assume that these remaining
non-sensitive attributes are commonly observable, obtainable, or reasonably
guessable in practical deployments. This is a standard assumption in attribute
inference attacks~\cite{fredrikson2014privacy,zhao2021feasibility,
mehnaz2022your,jayaraman2022attribute}. For example, $x_{-1}$ may include
coarse profile or demographic attributes that the user discloses to access the
service, as well as attributes visible from a public profile. Accordingly, we
consider a side information model in which the attacker knows $x_{-1}$ but does
not know $x_1$.

We also assume that the attacker can observe the user's released prediction.
The model output is shown to the user and may be (i) directly observed by a
nearby adversary, (ii) captured through screen sharing, screenshots, or
telemetry logs accessible to the adversary, or (iii) inferred from downstream
actions triggered by the decision (e.g., a recommended item being displayed or 
a service being offered or withheld). For example, an adversary may obtain a
user's recommendation result (e.g., whether insulin is recommended) from a
shared screen or from telemetry accessible to a third
party~\cite{pan2018panoptispy,ftc2023goodrx}, and combine it with publicly
available non-sensitive attributes to infer the unknown target sensitive
attribute.

Given $x_{-1}$ and a prediction $y=f(x_1,x_{-1})$, the attacker's goal 
is to infer the value of $x_1$ or a narrow plausible range. When the model's 
decision is closely associated with $x_1$ (e.g., recommending insulin
primarily based on BMI), observing $y$ can yield a tight inference interval
for $x_1$.

\paragraph{Scenario II: Model Inversion}
In this setting, the attacker aims to reconstruct inputs that the model 
consistently predicts as a chosen target class, thereby recovering 
representative samples from the training data distribution for~that~class.

Under the label-only constraint, the attacker iteratively queries the model and
updates a synthetic input based solely on observed label changes. For example,
in the label-only inversion attack of~\cite{kahla2022label} that we evaluate,
the attacker estimates a directional optimization signal by probing the model
with small perturbations around the current input and observing whether the
released label changes. The attacker's objective is to navigate the input space
toward regions where the model consistently predicts the target label. By
reaching such regions, the attacker recovers inputs that expose class-specific
features represented in the training data distribution.

\paragraph{Scenario III: Model Distillation as a Boundary Case}
We consider model distillation as a boundary case for RP. In this setting,
the attacker submits many inputs, collects the released labels, and trains a
student model to approximate the deployed model's global input--output behavior.

This scenario differs from Scenarios~I and~II in the protection target.
Scenarios~I and~II concern attribute-level and instance-level leakage: what a
released prediction reveals about a particular queried input's attribute, or how
released predictions guide reconstruction of representative samples from the
training data distribution. Model distillation, in contrast, is a function-level
extraction attack, which targets the model's global decision behavior aggregated
over many queries. We include Scenario~III to delineate the scope of RP rather
than as a primary threat that RP is intended to mitigate: RP bounds per-prediction 
leakage (Scenarios~I and~II) and is not intended to stop function-level extraction 
from aggregated query behavior.

%% file: privacy.tex
\section{Robust Privacy}
\label{sec:rp-def}

We formalize Robust Privacy (RP) as an inference-stage privacy
notion and Robust Attribute Privacy (RAP) as its attribute-level projection.
The certified robust radius $R$ serves as the privacy metric reported
throughout the paper.

\paragraph{Robust Privacy (RP)}
We first formalize Robust Privacy, which characterizes output invariance within
a certified neighborhood around an input as an inference-stage privacy~guarantee.

\begin{definition}[Robust Privacy (RP)]
\label{def:robust-privacy}
Let $f: \mathcal{X} \to \mathcal{Y}$ be a classifier, and let $\|\cdot\|_p$
be a specified $\ell_p$ norm. An input $x \in \mathcal{X}$ satisfies
$(R,\alpha)$-Robust Privacy under the released prediction of $f$ if, with
probability at least $1-\alpha$, the following invariance property holds:
for every $x' \in \mathcal{X}$ with
$\|x' - x\|_p < R$,
\[
f(x') = f(x).
\]
\end{definition}

Common choices of $\|\cdot\|_p$ include the $\ell_1$, $\ell_2$, and $\ell_\infty$
norms; unless otherwise specified, we use the $\ell_2$ norm in this study.
When RP is instantiated via deterministic verification (e.g., CROWN~\cite{crown},
$\beta$-CROWN~\cite{beta-crown}), the invariance guarantee holds deterministically,
i.e., with $\alpha=0$. In this study, we instantiate RP using randomized smoothing,
where the certified prediction invariance guarantee holds with failure probability
at most a user-specified $\alpha>0$. The following corollary shows that Gaussian
randomized smoothing provides a probabilistic instantiation of RP.

\begin{corollary}[RP instantiated with randomized smoothing]
\label{cor:rp-smoothing}
Let $h:\mathbb{R}^d\to\mathcal{Y}$ be a base classifier.
Let $g$ be the Gaussian smoothed classifier defined by
\[
g(x)
=
\arg\max_{c\in\mathcal{Y}}
\Pr_{\xi}\bigl[h(x+\xi)=c\bigr],
\qquad
\xi\sim\mathcal{N}(0,\sigma^2 I_d),
\]
where $\sigma>0$ is the noise scale. For an input $x\in\mathbb{R}^d$,
let $c_A=g(x)$ be the top class of the smoothed classifier. Suppose that,
with confidence at least $1-\alpha$, the class-probability bounds satisfy
\[
\Pr_{\xi}\bigl[h(x+\xi)=c_A\bigr] \ge \underline{p}_A
>
\overline{p}_B
\ge
\max_{c\neq c_A}
\Pr_{\xi}\bigl[h(x+\xi)=c\bigr].
\]
Then $x$ satisfies $(R,\alpha)$-Robust Privacy under the released prediction
of $g$ with respect to the $\ell_2$ norm, where
\[
R =
\frac{\sigma}{2}
\Bigl(
\Phi^{-1}(\underline{p}_A)
-
\Phi^{-1}(\overline{p}_B)
\Bigr),
\]
and $\Phi^{-1}$ denotes the inverse CDF of the standard Gaussian distribution.
\end{corollary}

\begin{proof}
By Theorem~1 of Cohen et al.~\cite{cohen2019certified}, the above probability
bounds certify that, with confidence at least $1-\alpha$, $g(x')=g(x)=c_A$ for
every $x'$ satisfying $\|x'-x\|_2<R$. This is exactly the invariance property
required by Definition~\ref{def:robust-privacy} for $(R,\alpha)$-Robust Privacy
under the released prediction~of~$g$.
\end{proof}

\paragraph{Privacy semantics of RP}
Beyond the geometric invariance stated in Definition~\ref{def:robust-privacy},
$(R,\alpha)$-Robust Privacy admits an information-theoretic indistinguishability
bound on any adversary that observes the released prediction through the
inference side channel. This makes precise the sense in which RP is a privacy
notion: within the certified radius $R$, the advantage of any adversary in
distinguishing the queried input from its neighbors through this side channel
is bounded by the certification failure probability $\alpha$. We denote this
certified neighborhood by
\[
\mathcal{B}_p(x,R)
\triangleq
\{x' \in \mathcal{X}: \|x'-x\|_p < R\}.
\]

\begin{theorem}[RP as inference side-channel indistinguishability]
\label{thm:rp-game}
Suppose $x\in\mathcal{X}$ satisfies $(R,\alpha)$-Robust Privacy under the
released prediction of $f$. Consider the following neighborhood-indistinguishability
game between a challenger and an adversary $\mathcal{A}$:
\begin{enumerate}
\item $\mathcal{A}$ selects any $x'\in\mathcal{B}_p(x,R)$;
\item the challenger samples $b\in\{0,1\}$ uniformly, sets $x_0=x$ and
$x_1=x'$, and releases $y=f(x_b)$;
\item $\mathcal{A}$ observes $y$ and outputs a guess $b'\in\{0,1\}$.
\end{enumerate}
Then for every adversary $\mathcal{A}$,
\[
\Bigl|\Pr[b'=b]\;-\;\tfrac{1}{2}\Bigr|
\;\le\;
\frac{\alpha}{2}.
\]
\end{theorem}

\begin{proof}
Let $E$ denote the event $\{f(x_0)=f(x_1)\}$, with probability taken over 
the randomness underlying the $(R,\alpha)$-Robust Privacy guarantee.
Since $x'\in\mathcal{B}_p(x,R)$, Definition~\ref{def:robust-privacy} gives 
$\Pr[E]\ge 1-\alpha$. Conditional on $E$, the released value $y=f(x_b)$ equals 
$f(x_0)$ regardless of $b$, so $\mathcal{A}$'s view is independent of $b$ and 
$\Pr[b'=b\mid E]=\tfrac{1}{2}$. Conditional on the complement, 
$\Pr[b'=b\mid\overline{E}]\le 1$ trivially. Combining,
\[
\Pr[b'=b]
\;\le\;
(1-\alpha)\cdot\tfrac{1}{2} \;+\; \alpha\cdot 1
\;=\;
\tfrac{1}{2}+\tfrac{\alpha}{2},
\]
and symmetrically $\Pr[b'=b]\ge \tfrac{1}{2}-\tfrac{\alpha}{2}$.
\end{proof}

Theorem~\ref{thm:rp-game} ties the certification failure probability $\alpha$
directly to an adversary's distinguishing advantage: the released prediction
$f(x)$ is, with advantage at most $\alpha/2$, useless for distinguishing $x$
from any input within the certified $\mathcal{B}_p(x,R)$. Conditional on the
high-probability event $E$, the indistinguishability is exact: among inputs
in the certified neighborhood, the inference side channel carries zero
information about which one was queried. In this sense, the certified radius
$R$ is the size of the neighborhood within which the released prediction
provably leaves the queried input information-theoretically hidden through
the inference side channel, and $\alpha$ controls how often this guarantee
can fail to hold. This is the formal privacy content that Theorem~\ref{thm:rp-to-rap} 
below then projects from the input level to the attribute level.

\paragraph{Robust Attribute Privacy (RAP)}
Building on RP, we formalize Robust Attribute Privacy, which characterizes
the set of sensitive-attribute values that remain compatible with a released
prediction when the adversary's side information $x_{-1}$ is fixed.

\begin{definition}[Robust Attribute Privacy (RAP)]
\label{def:rap}
Consider a classifier $f:\mathcal{X}\to\mathcal{Y}$ and an input
$x=(x_1,x_{-1})$, where $x_1\in\mathcal{Z}\subseteq\mathbb{R}$ denotes a
sensitive target attribute and $x_{-1}$ denotes all remaining attributes.
Here, $\mathcal{Z}$ denotes the feasible domain of the sensitive attribute
under the fixed context $x_{-1}$, so that $f(z,x_{-1})$ is well-defined for
all $z\in\mathcal{Z}$. Define the induced one-dimensional classifier
\[
f_{x_{-1}}(z) \triangleq f(z,x_{-1}).
\]
Given the released prediction $y=f(x)$, the RAP-compatible set is
\[
I_y^{\mathrm{RAP}}(x_{-1})
\triangleq
\{z\in\mathcal{Z}: f_{x_{-1}}(z)=y\}.
\]
If $I_y^{\mathrm{RAP}}(x_{-1})$ is an interval, we refer to it as the
RAP-compatible interval.
\end{definition}

For a fixed context $x_{-1}$, $I_y^{\mathrm{RAP}}(x_{-1})$ captures the
sensitive-attribute values that an attacker cannot rule out after observing
the released prediction $y$. All values in this set are compatible with the
same prediction when paired with the fixed side information $x_{-1}$.
Thus, a larger RAP-compatible set, or a longer RAP-compatible
interval, leaves the attacker with a wider plausible range for the 
sensitive attribute and therefore lowers attribute-inference precision.
This formulation assumes that $x_{-1}$ is fixed as attacker side information,
a standard assumption in attribute inference attacks~\cite{fredrikson2014privacy,
zhao2021feasibility,mehnaz2022your,jayaraman2022attribute}.

\paragraph{From RP to RAP}
RP and RAP are connected by a direct implication: an input-level robust radius
$R$ certifies that the $R$-neighborhood of the sensitive-attribute value lies
inside the RAP-compatible set. The input-level invariance of RP therefore
projects onto a certified lower bound on the attribute-level uncertainty,
controlled by~$R$.

\begin{theorem}[RP implies a certified RAP-compatible sub-interval]
\label{thm:rp-to-rap}
Let $f:\mathcal{X}\to\mathcal{Y}$ be a classifier and let $x=(x_1,x_{-1})$
with $x_1\in\mathcal{Z}\subseteq\mathbb{R}$. Suppose $x$ satisfies
$(R,\alpha)$-Robust Privacy under the released prediction of $f$. Then,
with confidence at least $1-\alpha$,
\[
\{z\in\mathcal{Z} : |z-x_1| < R\} \;\subseteq\; I_{f(x)}^{\mathrm{RAP}}(x_{-1}).
\]
In particular, when $(x_1 - R,\, x_1 + R)\subseteq\mathcal{Z}$,
$I_{f(x)}^{\mathrm{RAP}}(x_{-1})$ contains a sub-interval of length $2R$
centered at $x_1$.
\end{theorem}

\begin{proof}
For any $z\in\mathcal{Z}$ with $|z-x_1|<R$, the inputs $(z,x_{-1})$ and
$(x_1,x_{-1})$ differ only in the sensitive coordinate, so
$\|(z,x_{-1})-(x_1,x_{-1})\|_p = |z-x_1| < R$ under the $\ell_p$ norm fixed
in Definition~\ref{def:robust-privacy}. By Definition~\ref{def:robust-privacy},
with confidence at least $1-\alpha$, $f(z,x_{-1}) = f(x_1,x_{-1}) = f(x)$,
and therefore $z\in I_{f(x)}^{\mathrm{RAP}}(x_{-1})$.
\end{proof}

Theorem~\ref{thm:rp-to-rap} delivers this bound as a certified sub-interval
of the RAP-compatible set, without requiring the set to be globally connected: even if
$I_{f(x)}^{\mathrm{RAP}}(x_{-1})$ consists of multiple disjoint components in
general, RP guarantees that the component containing $x_1$ contains the local
neighborhood $\mathcal{Z}\cap(x_1 - R,\, x_1 + R)$, which has length $2R$
whenever this neighborhood lies in the interior of $\mathcal{Z}$. The empirical
RAP-compatible interval lengths reported in Section~\ref{sec:moti} measure the
actual compatible region to which the attacker can, at best, localize the
sensitive value.

\paragraph{Privacy Metric}
For a fixed failure probability $\alpha$ and a specified norm $\|\cdot\|_p$, we
refer to any certified value of $R$ for which Definition~\ref{def:robust-privacy}
holds as a robust radius of input $x$ under the released prediction of the
classifier. Corollary~\ref{cor:rp-smoothing} gives one such radius for RP under
Gaussian smoothing.

We use the robust radius $R$ as the privacy metric in this study. At the input
level, the certified neighborhood $\mathcal{B}_p(x,R)$ is mapped to the same
released prediction as $x$.
Equivalently, $\mathcal{B}_p(x,R) \subseteq f^{-1}(f(x))$. Thus, by
Theorem~\ref{thm:rp-game}, an adversary observing the released prediction
$f(x)$ has advantage at most $\alpha/2$ (e.g., $\alpha=0.01$ in our
experiments) in distinguishing $x$ from other inputs within this
neighborhood. A larger $R$ therefore means that the released prediction
reveals less fine-grained information about the input. At the attribute
level, by Theorem~\ref{thm:rp-to-rap}, the same
$R$ serves as a certified lower bound: the RAP-compatible region
around $x_1$ contains the local neighborhood $\mathcal{Z}\cap(x_1 - R,\, x_1 + R)$
(of length $2R$ when this neighborhood lies in the interior of $\mathcal{Z}$), so
a larger $R$ directly enlarges the certified lower bound on the attribute-level
uncertainty the attacker faces under the released prediction.

\paragraph{Operational meaning of $R$ across input spaces}
The robust radius $R$ is a geometric privacy quantity, and its operational
content depends on what the coordinates of $\mathcal{X}$ mean. In the tabular
attribute-inference setting of Section~\ref{sec:moti}, where the sensitive
coordinate is a single continuous attribute on a known scale, $R$ admits an
immediate attribute-level reading via Theorem~\ref{thm:rp-to-rap}: the
released prediction cannot resolve the sensitive value below a certified
local neighborhood of length $2R$ around it. In the image setting of
Section~\ref{sec:instance-exp}, the $\ell_2$-ball $\mathcal{B}_2(x,R)$ is a
high-dimensional set of nearby candidate inputs that all map to
the same released label; the released label therefore cannot single out $x$
from this candidate set, and an inversion adversary that probes local label
changes around a synthetic input loses the directional signal it would
otherwise use to localize $x$ within the set. The same certificate $R$ thus
lower-bounds the attribute-level ambiguity in the first setting and the
instance-level ambiguity in the second.

%% file: exp_attribute.tex
\section{Experiments: Attribute-Level Privacy Protection against Attribute Inference}
\label{sec:moti}

This section gives the first empirical test of Robust Privacy (RP) and its
attribute-level projection, Robust Attribute Privacy (RAP). We focus on the
basic inference-interface concern raised by personalized model outputs: a 
classifier releases one hard label per query; a continuous sensitive attribute
of the queried input strongly influences that label; how much can the released
label tell an outside observer about that attribute, and how much of that
side-channel signal does RP suppress?

Since the sensitive attribute we track is continuous, the side-channel
leakage through the inference interface is measured as the length of the
interval of attribute values that remain compatible with the released label
given the adversary's side information. This attribute-level privacy measurement 
aligns the experiment with the RAP definition (Definition~\ref{def:rap}).

\paragraph{Experimental setup}
We use the Medical Insurance Cost Prediction
dataset~\cite{MedicalInsuranceCostPrediction2025}, which contains $100{,}000$
records with demographic, lifestyle, clinical, utilization, and insurance plan
features. We define a binary classification task from the high-risk indicator
provided by the dataset (\texttt{is\_high\_risk}), interpreted as a high-risk
care management label. To avoid label leakage, we remove the related
\texttt{risk\_score} and post-outcome insurance/claim fields from the model
inputs. The remaining inputs are continuous or ordinal health features
(age, BMI, blood pressure, lab measurements, medication count, hospitalization
history, and chronic-condition count), together with one-hot encoded categorical
features. The positive rate of \texttt{is\_high\_risk} is $36.78\%$. We use a
fixed stratified $60/40$ train/test split.

The base classifier is a GBNet~\cite{horrell2025gbnet} implementation of
LightGBM, trained on standardized features using an AMD EPYC 7302 16-core CPU,
with $200$ boosting rounds, maximum tree depth $4$, and at most $31$ leaves
per tree. The trained model achieves $100.00\%$ accuracy and AUC $1.00$ on
the held-out test split, making the released label highly informative
about the input under this task.

On top of this base classifier, we instantiate RP via randomized
smoothing~\cite{cohen2019certified}. For each input $x$, we certify the
released prediction of the smoothed classifier. The certification uses
$N\in\{1000,5000\}$ Monte Carlo samples, failure probability $\alpha=0.01$, 
and noise scales $\sigma\in\{0.1,0.2,\ldots,1.0\}$. 

\paragraph{RAP evaluation}
We use age as the sensitive target attribute. The attacker, as defined in
Scenario~I of Section~\ref{sec:threat}, knows the non-age attributes
$x_{-1}$ and observes the released label. In this task, a positive label
typically corresponds to a relatively larger age, so each positive released
label narrows the set of age values compatible with that label. The attacker
therefore tries to localize the RAP-compatible age range while holding
$x_{-1}$ fixed. We first examine the theoretical protection provided by RP in
this setting: the robust radius $R$ certified for each query projects to a
certified RAP-compatible sub-interval, as proved in Theorem~\ref{thm:rp-to-rap}. 
Under our privacy metric, a larger $R$ indicates better privacy protection 
for the target attribute, or less precise attribute inference. We then 
corroborate this theoretical implication with the following attribute-inference 
attack under the same target attribute and RP setting.

\begin{figure}[t]
    \centering
    \includegraphics[width=\linewidth]{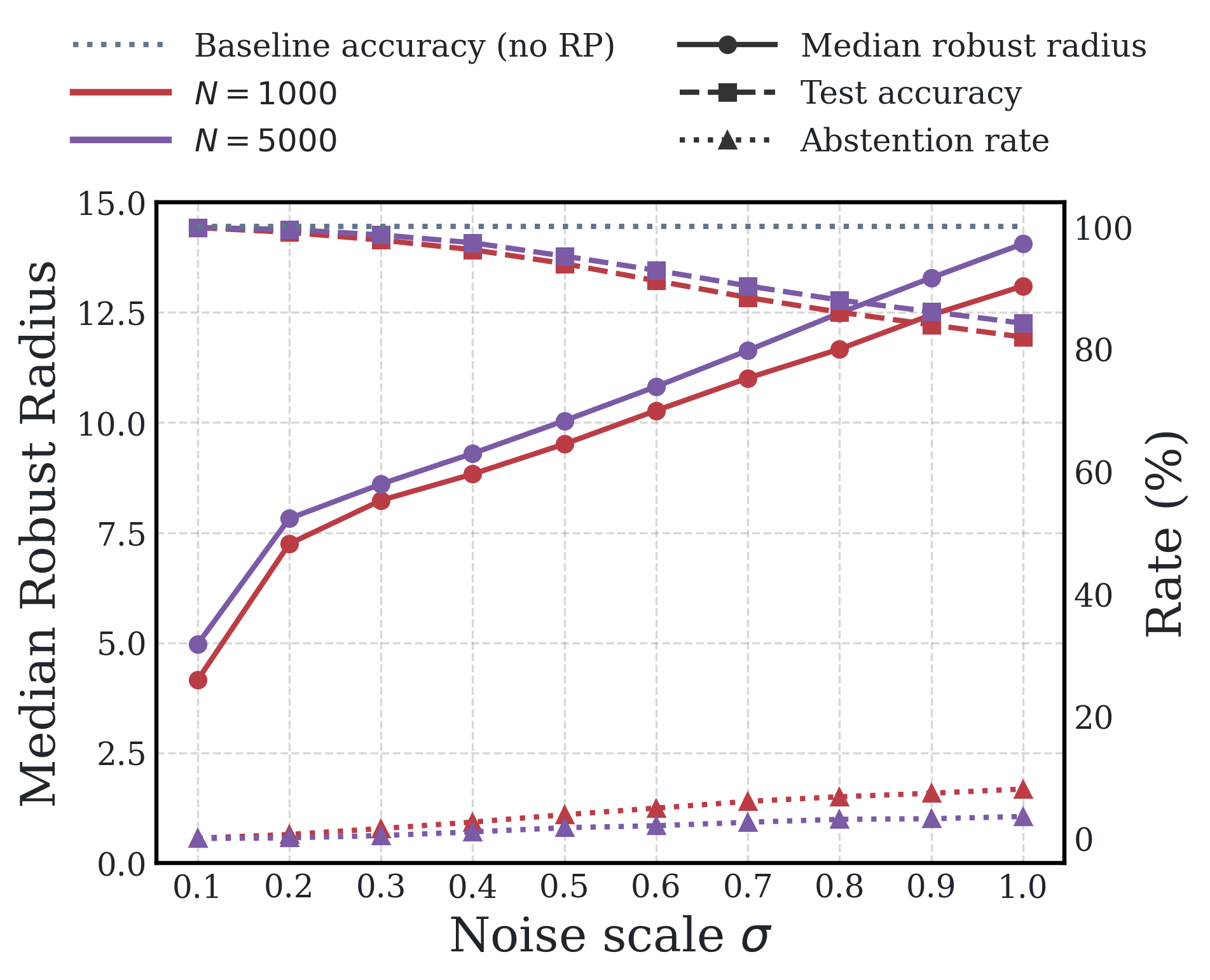}
    \caption{
        Median certified robust radius, test accuracy, and abstention rate for
        the high-risk care management classification task under RP, using
        $N\in\{1000,5000\}$ Monte Carlo samples, failure probability $\alpha=0.01$,
        and noise scales $\sigma\in\{0.1,0.2,\ldots,1.0\}$. Increasing $\sigma$
        yields the privacy--utility tradeoff: the median certified radius 
        rises while test accuracy decreases and the abstention
        rate increases. Increasing $N$ at fixed $\sigma$ behaves differently: the
        certified radius rises, test accuracy improves, and the abstention rate
        drops simultaneously, so $N$ improves both privacy and utility together
        rather than trading one for the other. If a deployer can afford higher
        per-query inference cost, raising $N$ therefore offers a concrete way
        to obtain stronger privacy and higher utility together.
    }
    \label{fig:rec-privacy-utility}
\end{figure}

Figure~\ref{fig:rec-privacy-utility} shows that the certified robust radius grows
smoothly from $4.15$ at $\sigma=0.1$ to $13.10$ at $\sigma=1.0$ when
$N=1000$, and from $4.97$ to $14.06$ when $N=5000$. The corresponding test
accuracy decreases from $99.84\%$ to $81.95\%$ at $N=1000$ and from $99.86\%$ to
$84.21\%$ at $N=5000$, while the abstention rate increases from $0.09\%$ to
$8.12\%$ at $N=1000$ and from $0.03\%$ to $3.63\%$ at $N=5000$. 
Increasing $\sigma$ produces the expected privacy--utility tradeoff: stronger
smoothing gives a larger certified invariance region around each queried input,
which corresponds to stronger inference-stage privacy but also lower classification
accuracy. Increasing $N$ at fixed $\sigma$ behaves differently: the certified
radius rises, test accuracy improves, and the abstention rate drops simultaneously,
so $N$ improves both privacy and utility together rather than trading one for
the other. If a deployer can afford higher per-query inference cost, raising $N$
therefore offers a concrete way to obtain stronger privacy and higher utility together.

\paragraph{Attribute inference precision under RP}

Under the same RP setting as the certified robust radius evaluation above 
(i.e., the same base classifier, $N\in\{1000,5000\}$, $\alpha=0.01$, and
$\sigma\in\{0.1,0.2,\ldots,1.0\}$), we now validate that increases in the
robust radius $R$ translate into larger empirical RAP-compatible intervals 
for the sensitive attribute, thereby reducing attribute-inference precision. 
We conduct an attribute-inference attack on age. For a test input
$x=(\mathrm{age},x_{-1})$ with a positive released prediction, the attack 
keeps all non-age attributes $x_{-1}$ fixed, queries the model with 
$(z,x_{-1})$ for candidate age values $z$, and records whether the released 
prediction remains positive.

The attack starts the search for the plausible age interval with $[0,60]$,
given the positive released label and the fixed $x_{-1}$. The right endpoint
$60$ is the fixed high-risk reference endpoint, consistent with a common
care-management convention that age $\geq 60$ marks elevated chronic-disease
risk. We fix this right endpoint because the metric measures how far the 
positive label remains compatible when moving leftward from a high-risk age; 
searching to the right would not flip a positive prediction into a negative 
one in this setting, and therefore would not help identify the compatible 
interval. Because the right endpoint is fixed, narrowing the plausible age 
interval reduces to localizing the left boundary $\mathrm{age}_{\mathrm{left}}$ 
of the positive prediction region along the age coordinate. The attack first 
checks whether the prediction at age $60$ is positive. If not, the sample is 
discarded because no positive label can be produced in the interval anchored 
at this reference endpoint. The attack then checks whether the prediction at 
age $0$ is positive. If so, the attack stops and $\mathrm{age}_{\mathrm{left}}$ 
is set to $0$. Otherwise, the attack performs binary search between $0$ and $60$ 
until the boundary is localized to within $0.01$ years, using at most $100$
queries per sample. For each kept sample, the compatible interval length
is then $60 - \mathrm{age}_{\mathrm{left}}$.

\begin{figure*}[t]
    \centering
    \includegraphics[width=\textwidth]{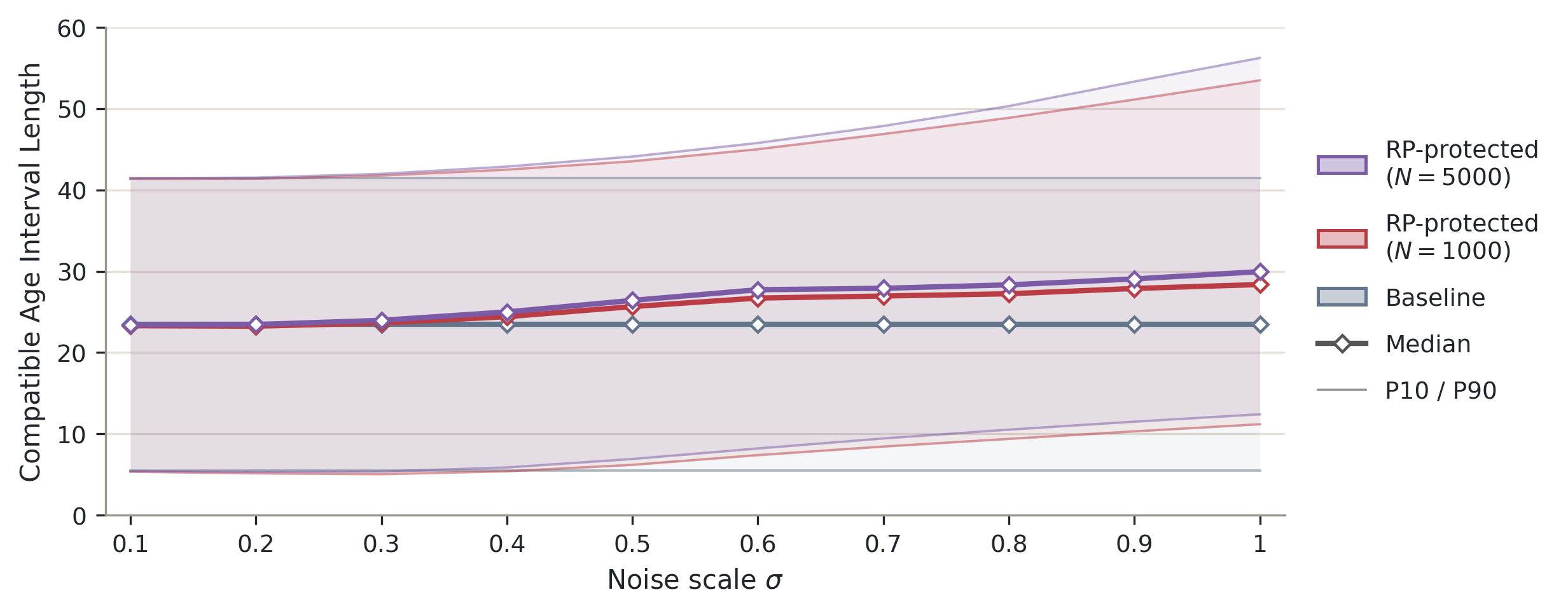}
    \caption{
        Distribution of RAP-compatible interval lengths over
        $\sigma\in\{0.1,0.2,\ldots,1.0\}$, with $\alpha=0.01$ and $N\in\{1000,5000\}$.
        Thick lines with diamond markers show the median compatible interval
        length at each $\sigma$; thin lines show the $10$th and $90$th
        percentiles, and the shaded band between them shows where the middle
        $80\%$ of test samples fall. Larger compatible interval lengths mean 
        that the attacker can only constrain the sensitive age attribute to a 
        wider interval, i.e., the label reveals less precise age information. 
        As $\sigma$ increases, RP more clearly expands the compatible age interval 
        relative to the unprotected baseline. Increasing $N$ at fixed $\sigma$ 
        further widens the RAP-compatible interval, with the test accuracy
        also rising, suggesting that RP mitigates attribute inference by
        suppressing fine-grained leakage at the inference interface, rather
        than by degrading model performance. If a deployer can afford higher 
        per-query inference cost, raising $N$ therefore offers a concrete way 
        to lower attribute-inference attack precision while also raising test accuracy.
    }
    \label{fig:rec-rap-distribution}
\end{figure*}

A larger RAP-compatible age interval means less precise age inference and
stronger protection. Hence, we compare the length of the compatible age interval
for a positive prediction without and with RP on the same samples. We refer to
these two cases as the baseline and RP-protected cases, respectively. For each
$\sigma$, we select test samples for which both the base classifier and the 
RP-protected smoothed classifier predict positive, thereby comparing the 
distribution of compatible interval lengths on the same sample set. 
Figure~\ref{fig:rec-rap-distribution} shows this~comparison.

Figure~\ref{fig:rec-rap-distribution} shows how the certified robust radius
translates into empirical expansion of the RAP-compatible interval. The
expansion of the RAP-compatible age-inference interval under RP becomes clearer
as $\sigma$ increases, as shown by the diverging median compatible interval
lengths of the baseline and RP-protected cases. At $\sigma=0.6$, RP expands the
median compatible interval length to $26.70$ years at $N=1000$ and to $27.73$
years at $N=5000$, while retaining $91.15\%$ and $92.86\%$ test accuracy
respectively; at $\sigma=1.0$, the median compatible interval length reaches
$28.37$ years at $N=1000$ and $29.96$ years at $N=5000$, with test accuracy
$81.95\%$ and $84.21\%$ respectively. In contrast,
the median compatible interval length in the baseline case stays almost
unchanged at $23.50$ years. Holding $\sigma$ fixed, raising $N$ from $1000$
to $5000$ increases both the median RAP-compatible interval length and the
test accuracy, so raising $N$ incurs no privacy--utility trade-off: it
improves both at once. We observe a similar effect in the inversion 
experiment in Section~\ref{sec:instance-exp}, where raising $N$ at fixed 
$\sigma$ also lowers attack success rate while improving accuracy.

The certified robust radius and the empirical compatible interval length
are measured under the same RP setting. Under this shared setting, 
they measure certified and empirical quantities, respectively: 
the radius $R$ is a defender-side certified lower bound on label invariance, 
while the compatible interval length is an attacker-side empirical measurement 
of attribute uncertainty after observing the label. Theorem~\ref{thm:rp-to-rap} 
ties the two together: the certified radius $R$ guarantees a corresponding 
RAP-compatible neighborhood around the true age value, so the empirical 
compatible interval lengths reported in Figure~\ref{fig:rec-rap-distribution} 
should grow with $R$. The measurements bear this out: at $N=5000$, as $\sigma$
increases from $0.1$ to $1.0$, the certified radius $R$ grows from $4.97$ to
$14.06$ and the median compatible interval length increases from $23.42$
to $29.96$, indicating that RP delivers the formally guaranteed
attribute-level privacy protection in practice. 

We note that the measured expansion reported above is smaller than the
growth of the certified diameter $2R$. Theorem~\ref{thm:rp-to-rap} accounts for this: 
the full length-$2R$ sub-interval is guaranteed only when 
$(x_1 - R,\, x_1 + R)\subseteq\mathcal{Z}$, and the attack instantiates 
$\mathcal{Z}$ as the bounded search range $[0,60]$, which does not accommodate 
all attribute values compatible with the positive label. The part of the 
positive-label compatible region falling outside this range does not
enter the measured interval length, so the measured expansion understates
the growth of the certified guarantee.

%% file: exp_instance.tex
\section{Experiments: Instance-Level Privacy Protection against Model Inversion}
\label{sec:instance-exp}

This section anchors RP empirically at the instance level. Section~\ref{sec:moti}
already evaluated RP against attribute-level inference from a single released
prediction; here we test whether the indistinguishable neighborhood established
around each individual query also suppresses an instance-level threat that probes
the inference interface across many~queries.

The threat we study is model inversion, which reconstructs a representative
input for a target class by iteratively refining a synthetic input using
prediction feedback from the deployed model. We evaluate a label-only black-box
attack, which estimates update directions from how the released prediction
changes under small probes around the current synthetic input. The attack does
not require confidence scores, training data access, or any auxiliary signal
beyond the inference interface itself. Its attack signal lives entirely in the
inference interface of the deployed model.

\begin{figure}[h]
    \centering
    \includegraphics[width=\linewidth]{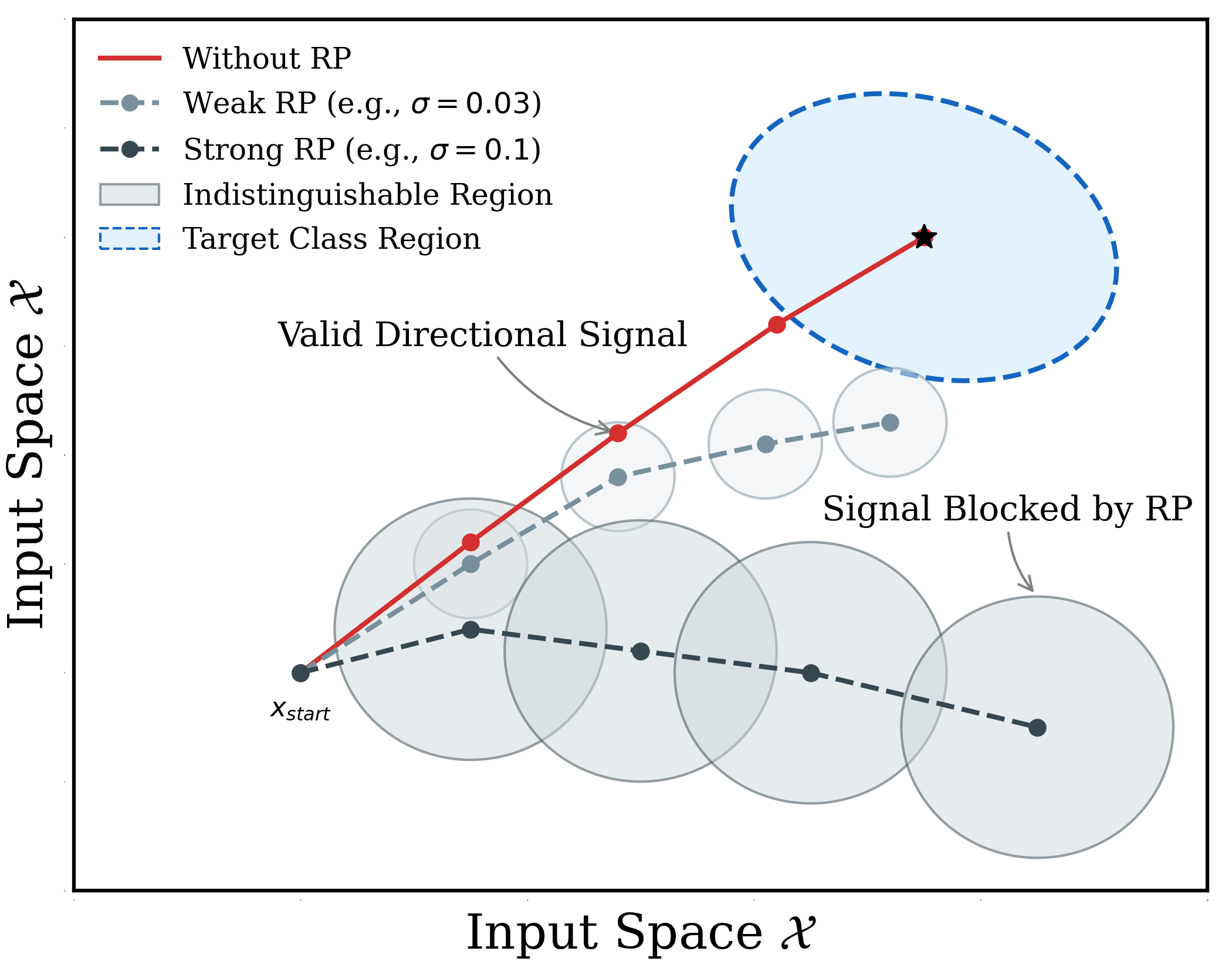}
    \caption{
        Mechanism of RP for mitigating model inversion attacks. Without RP
        (red solid line), the attacker exploits local prediction changes to
        estimate an optimization direction and steer updates toward the target
        region. With RP (dashed lines), predictions remain invariant within a
        robust radius $R$, masking the optimization signal used for iterative
        reconstruction toward the target region.
    }
    \label{fig:rp_mia_logic}
\end{figure}

This is exactly the side-channel leakage that RP is designed to suppress.
As illustrated in Figure~\ref{fig:rp_mia_logic}, once the released prediction
is certified to be invariant within an $R$-bounded neighborhood of the queried
input, the probes that the attack relies on no longer return informative
changes inside that neighborhood: there is no reliable local directional signal
for the attacker to read, and the optimization loses the cue it was built on.
The instance-level question this section asks is therefore not only whether RP
mitigates the attack at all---the geometry suggests that it should---but how
the privacy--utility tradeoff achieved by RP compares to alternative defenses
at the inference and training stages, such as randomized response and DP-SGD.

\paragraph{Baselines}
We compare RP with three baselines. The first is the unprotected base 
classifier, which uses the same target model and the same label-only 
attack interface but applies no privacy protection. This baseline isolates 
the effect of the protection mechanism: any reduction in attack success 
relative to this case can be attributed to the added protection.

The second and third baselines are privacy-preserving machine learning
mechanisms discussed in Section~\ref{sec:related:ppml} that fit the label-only
inference interface attack setting. The second is DP-SGD~\cite{abadi2016deep}, 
a training-stage privacy protection method that clips per-example gradients and 
adds Gaussian noise when training the target classifier. DP-SGD is a natural 
comparison because it is a gold standard for protecting training records. DP-SGD 
and RP mitigate training-data leakage at different stages: DP-SGD reduces memorization
of training records during training, while RP masks the fine-grained signals of
input--output dependence leaked through the inference interface. We should note
that, no matter where training data protection is enforced, attacks targeting
training data are typically conducted by exploiting privacy leakage through the
inference interface. We include this baseline to show that, when privacy leakage
happens at the inference stage, it can be more effective to enforce protection
at the inference~stage.

The third baseline is randomized response~\cite{warner1965randomized,
opendp_randomized_response}, a hard-label randomization baseline. After the base
classifier predicts a label, the API releases that label with probability $p$
and otherwise releases a randomly chosen alternative class. This baseline is
well matched to the attack interface: since the adversary only observes hard
labels, randomized response directly perturbs the released prediction as a side
channel. We include this baseline to show that, compared to randomized response,
RP instantiated with randomized smoothing can preserve substantially more utility
while mitigating privacy leakage through the inference interface, because its
final prediction is stabilized by aggregation over Gaussian-perturbed copies of
the queried input.

\paragraph{Experimental setup}
We use randomized smoothing to instantiate RP. The protected model
returns a label for every query by majority vote over noisy evaluations,
without checking certification confidence or returning an abstention. We use this
always-return-a-label protocol to avoid a trivial defense in which the attack fails
simply because the model refuses to answer. Thus, reductions in attack success
reflect RP's masking of fine-grained input--output dependence signals rather than
query rejection. We run the experiment on NVIDIA A100-SXM4-40GB GPUs.

We use the model inversion attack of Kahla et al.~\cite{kahla2022label} and
follow their recommended setting. The target model is FaceNet64, and a separate
FaceNet~\cite{schroff2015facenet} classifier is used as the evaluator. Both
models classify the same $1000$ private identities from CelebA~\cite{liu2015deep},
but use different architectures so that attack success reflects semantic recovery
rather than overfitting to a particular evaluator. We attack target classes
$0$--$99$ and evaluate prediction accuracy on a fixed set of $1000$ CelebA
images, one per private identity.

We evaluate the effectiveness of RP in mitigating the model inversion attack
across $\sigma\in\{0.01,0.02,\ldots,0.10\}$ and $N\in\{10,100\}$. For DP-SGD, we train
FaceNet64 with PyTorch's Opacus~\cite{opacus_api_reference}, replacing batch
normalization layers with group normalization to support per-example gradient
accounting. We use SGD for $50$ epochs with batch size $64$, learning rate
$0.01$, momentum $0.9$, weight decay $10^{-4}$, clipping norm $10$, and noise
multipliers $\sigma_{\mathrm{SGD}}\in\{0.01,0.02,\ldots,0.05\}$. For randomized 
response, we evaluate its mitigation effectiveness with keep probabilities 
$p\in\{0.65, 0.67, 0.70, 0.72, 0.75, 0.77, 0.85, 0.95\}$.

\begin{figure}[h]
    \centering
    \includegraphics[width=\linewidth]{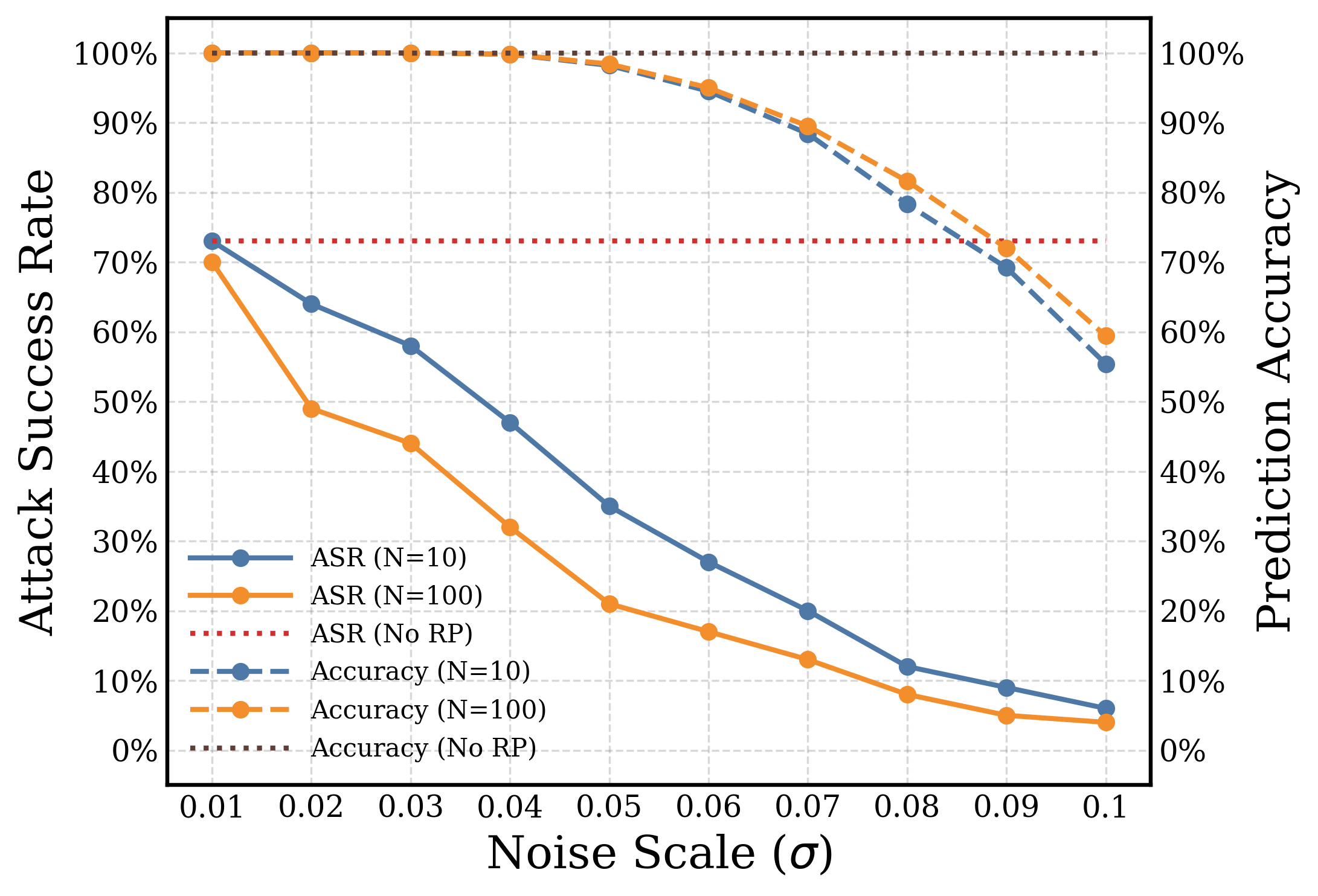}
    \caption{
        Model inversion under a label-only, black-box setting~\cite{kahla2022label}:
        attack success rate (ASR; solid lines, left y-axis) and prediction accuracy
        (dashed lines, right y-axis) for the base classifier and smoothed classifiers
        across noise scales $\sigma\in\{0.01,0.02,\ldots,0.10\}$ with $N\in\{10,100\}$.
        Horizontal dotted lines denote the no-RP baselines (ASR $=73\%$ and
        accuracy $=100\%$). Enabling RP reduces ASR monotonically as $\sigma$
        increases. Increasing $N$ increases inference cost but can improve utility
        while further lowering ASR, suggesting that RP mitigates model inversion by
        suppressing fine-grained privacy leakage through the inference interface
        rather than by degrading model performance.
    }
    \label{fig:attack-accuracy-noise}
\end{figure}

\paragraph{Results}
Figure~\ref{fig:attack-accuracy-noise} reports attack success rate (ASR) and
model accuracy. Without RP, the attack achieves $73\%$ ASR. RP reduces ASR as
$\sigma$ increases: at $\sigma=0.1$ and $N=100$, ASR drops to $4\%$, while the
smoothed classifier still maintains non-trivial accuracy. At $\sigma=0.03$ and
$N=100$, accuracy remains $100\%$ while ASR drops to $44\%$, demonstrating
partial mitigation without model performance degradation. For a fixed $\sigma$,
increasing $N$ from $10$ to $100$ improves utility while further lowering ASR.
We see a similar phenomenon in the attribute-inference experiment in
Section~\ref{sec:moti}: at fixed $\sigma$, raising $N$ from $1000$ to
$5000$ both lowers attribute-inference precision and raises test
accuracy. Across the two experiments, this suggests that RP mitigates the 
inversion attack by suppressing fine-grained privacy leakage through the inference 
interface rather than by degrading the model performance.

\begin{figure}[h]
    \centering
    \includegraphics[width=\linewidth]{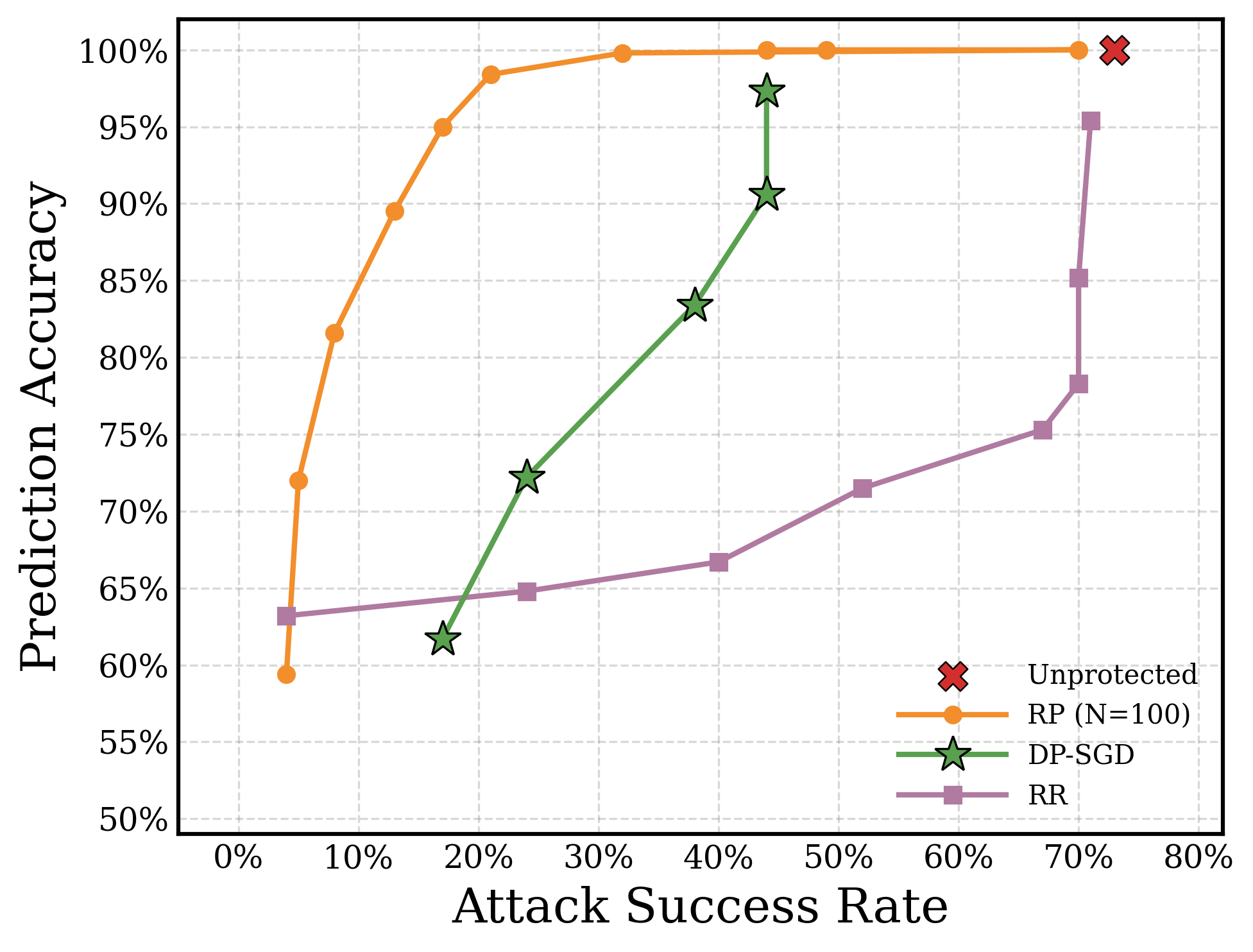}
    \caption{
        Privacy--utility tradeoff of RP compared with DP-SGD and randomized
        response (RR). The x-axis is attack success rate (ASR), and the y-axis
        is prediction accuracy; better privacy--utility points lie toward the
        upper left of the figure, where ASR is lower and accuracy is higher.
        For RP, $\sigma$ varies over $\{0.01,0.02,\ldots,0.10\}$ with $N=100$. For DP-SGD, we
        train FaceNet64 with noise multipliers $\sigma_{\mathrm{SGD}}\in\{0.01,0.02,\ldots,0.05\}$. 
        For RR, we evaluate keep probabilities $p\in\{0.65, 0.67, 0.70, 0.72, 0.75, 0.77, 0.85, 0.95\}$. The RP points
        lie closer to the upper left of the figure, showing that RP achieves an 
        overall better privacy--utility tradeoff than DP-SGD and RR. At high 
        utility, RP with $\sigma=0.05$ keeps $98.4\%$ accuracy at $21\%$ ASR, 
        whereas the best DP-SGD point at high utility ($\sigma_{\mathrm{SGD}}=0.01$)
        keeps $97.3\%$ accuracy at $44\%$ ASR, and RR with $p=0.95$ keeps
        $95.4\%$ accuracy at $71\%$ ASR.
    }
    \label{fig:inversion-baseline-tradeoff}
\end{figure}

Figure~\ref{fig:inversion-baseline-tradeoff} compares RP ($N=100$) with DP-SGD
and randomized response under the same attack setting. The horizontal and vertical 
axes correspond to attack success rate and prediction accuracy, respectively. 
Therefore, better privacy--utility points lie toward the upper left of the figure, 
where ASR is lower and accuracy is higher. As shown, RP achieves a better 
privacy--utility tradeoff than DP-SGD and randomized response. At high utility, 
RP with $N=100$ and $\sigma=0.05$ keeps $98.4\%$ accuracy while reducing ASR to 
$21\%$. In contrast, the best high utility DP-SGD point keeps $97.3\%$ accuracy but
leaves ASR at $44\%$, and randomized response with $95.4\%$ accuracy leaves
ASR at $71\%$, close to the unprotected baseline. DP-SGD can reduce ASR further
only when utility drops substantially; for example, at $61.7\%$ accuracy, it
reaches $17\%$ ASR. RP reaches the same $17\%$ ASR at $95.0\%$ accuracy
($\sigma=0.06$, $N=100$).

The difference in the privacy--utility tradeoff follows from what each baseline
perturbs. Randomized response randomly corrupts the released label, while RP
stabilizes its prediction by aggregating noisy predictions over Gaussian-perturbed 
copies of the queried input. Hence, at comparable mitigation effectiveness 
(or comparable ASR), RP preserves substantially more model utility than randomized 
response. DP-SGD reduces training-stage memorization, while the attack signal 
used by model inversion is exposed at the inference stage. This creates a mismatch 
between the protection locus and the leakage channel: DP-SGD reduces one source 
of training-data leakage during training, but does not directly suppress the 
inference-interface side-channel signal exploited by the attack. RP instead targets 
this inference-stage privacy leakage directly: it establishes an indistinguishable 
neighborhood around the queried input under the released prediction, masking the 
fine-grained signals of input--output dependence that model inversion relies on 
to optimize its synthetic inputs toward the training data distribution of the 
target class.

We should also note the privacy budgets at which DP-SGD is run. 
As $\sigma_{\mathrm{SGD}}$ decreases from $0.05$ to $0.01$ with
$\delta=10^{-5}$, the corresponding privacy budget $\varepsilon$ grows from
$3.25\times 10^{6}$ to $1.15\times 10^{8}$. At these magnitudes, the formal
DP guarantee no longer carries quantitative meaning: the $\varepsilon$ values
are far above the regime in which DP composition theorems give a non-vacuous
bound on individual-record leakage. This is not a tuning artifact specific 
to our setup. It is a recurring observation in the DP deep learning
literature~\cite{tramer2021dp,de2022unlocking,carlini2022membership} that
training a moderately sized neural network at small $\varepsilon$ can collapse
utility well below what a deployed system would tolerate, and that
utility-matched comparisons against non-DP defenses therefore tend to live 
in this large-$\varepsilon$ regime. We argue that this is precisely the
predicament that motivates an inference-stage privacy defense. When a
training-stage mechanism cannot simultaneously deliver deployable utility and a
quantitatively meaningful $\varepsilon$ on tasks of this scale, the privacy
burden for the inference interface side channel has to be carried somewhere
else. Using stronger DP noise would mainly move the DP-SGD baseline along the
utility-degradation direction; it would not directly suppress the
inference-stage input--output dependence signal exploited by model inversion.
The gap is structural, and it reflects a mismatch between the protection 
locus (training-stage memorization) and the leakage channel (inference-stage
input--output dependence).

%% file: exp_distillation.tex
\section{Experiments: Function-Level Model Distillation as a Boundary Case}
\label{sec:distillation}

The classification and model inversion experiments evaluate RP in the
attribute-level and instance-level settings it is designed to protect.
In this section, we examine a different level of threat: model distillation
(or model extraction). In this attack, the adversary queries the deployed model
with many inputs, collects input--output pairs, and trains a student model
to imitate the target model's global input--output behavior, i.e., the model's
capability. As discussed above, RP mitigates privacy leakage through the
inference interface for individual queried inputs. We now use model distillation
as a boundary case to examine whether this per-query privacy protection extends
to an attack that learns the target model's global behavior from many queries
(i.e., a function-level~extraction~attack).

\paragraph{Experimental setup}
We evaluate the effectiveness of RP against model distillation using
DisGUIDE~\cite{rosenthal2023disguide}, a distillation attack that
requires only hard-label feedback. The teacher is a ResNet34-8x model
on the CIFAR-10 dataset. The adversary trains a distilled student from
input--output pairs collected by querying the teacher through the inference
interface. We apply RP, instantiated with randomized smoothing, to the 
teacher across $\sigma\in\{0.01,0.02,\ldots,0.10\}$ and $N\in\{10,100\}$, using the same
always-return-a-label protocol as in Section~\ref{sec:instance-exp}. We 
report teacher accuracy as utility and distilled model accuracy as attack
effectiveness. The experiment is conducted on an NVIDIA A100-SXM4-40GB~GPU.

\begin{figure}[t]
    \centering
    \includegraphics[width=\linewidth]{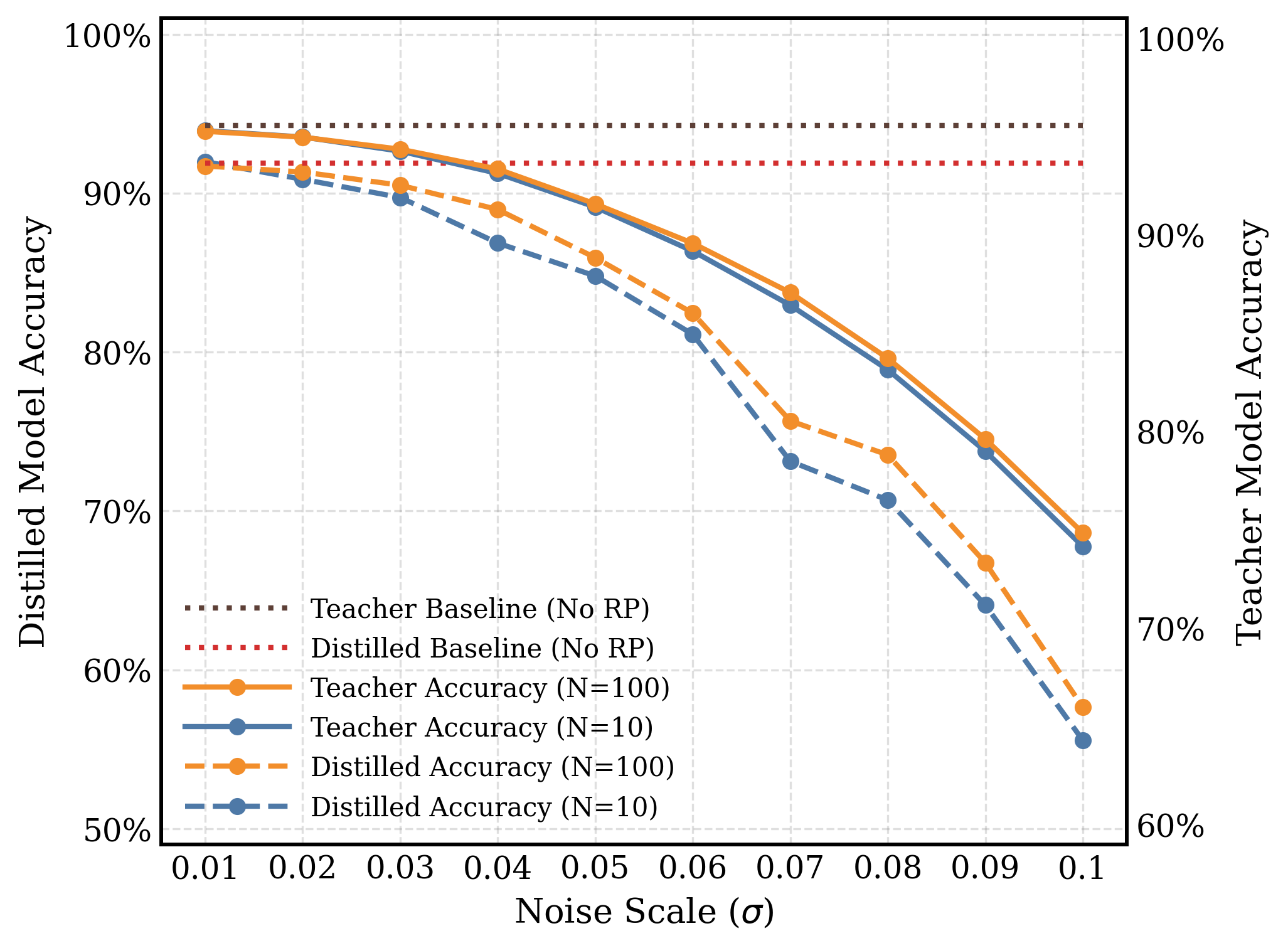}
    \caption{
        DisGUIDE~\cite{rosenthal2023disguide} model distillation against a
        ResNet34-8x model trained on the CIFAR-10 dataset under a label-only,
        black-box setting. Solid lines report RP-protected teacher accuracy;
        dashed lines report distilled model accuracy, which measures attack
        effectiveness. Horizontal dotted lines denote the test accuracy of 
        the no-RP teacher and the corresponding distilled model baselines.
        Increasing $\sigma$ reduces both teacher accuracy and distilled model
        accuracy. At fixed $\sigma$, decreasing $N$ degrades teacher accuracy 
        and also lowers distilled model accuracy, showing that the relationship 
        between target model utility and attack effectiveness differs from that 
        in the attribute-inference and model inversion experiments. We attribute 
        this pattern to the function-level attack surface of model distillation,
        compared with the attribute-level and instance-level attack surfaces
        of attribute inference and model inversion. In this setting, the
        mitigation of model distillation at least partly originates from
        target model utility degradation.
    }
    \label{fig:distillation-attack-accuracy-noise}
\end{figure}

\paragraph{Results}
Figure~\ref{fig:distillation-attack-accuracy-noise} shows a relationship 
between target model utility and attack effectiveness that differs from 
the attribute-inference and model inversion experiments. As $\sigma$ 
increases, teacher accuracy decreases, and distilled model accuracy, 
which measures attack effectiveness, also decreases with it. At fixed 
$\sigma$, decreasing $N$ from $100$ to $10$ lowers the RP-protected 
teacher's accuracy and also tends to lower the distilled student's 
accuracy. Thus, unlike in the attribute-inference and model inversion 
experiments, where raising $N$ at fixed $\sigma$ improved utility and 
privacy at once, the mitigation of model distillation seems to be partly 
driven by target model utility degradation.

We attribute this change in the relationship between target model utility and
attack effectiveness to the different level of the attack surface. Distillation
is a function-level extraction attack: it does not rely on inferring a specific
attribute or reconstructing a specific instance. Instead, it learns a global
approximation of the deployed model. RP certifies local prediction invariance
for each queried input, so it can mask the fine-grained per-query input--output
dependence signals used by attribute-level and instance-level attacks, but the
coarser global input--output relationship can still be learned from many
queries.

This clarifies the scope of RP: RP mitigates per-query information leakage 
(e.g., attribute-level and instance-level) through the inference interface, 
but it is not designed to prevent global function-level extraction.

%% file: discuss.tex
\section{Discussion}
\label{sec:discus}

We discuss interpretation issues for RP, including its intuition, scope, 
and relationship to DP and certified robustness.

\subsection{Intuition Behind Robust Privacy}
\label{sec:sub:rp_intuition}

There is a natural way to picture what RP does, and we believe it explains
why the mechanism behaves the way it does in our experiments.

Imagine the input space partitioned by the classifier into colored regions,
one color per class. A queried input is a point inside one such region.
Without protection, the released label tells the outside observer which region
the point sits in, but the observer can do much more than that: by probing
nearby points, the observer can also trace where the boundary of the region
runs near the queried point, and from the local shape of that boundary infer
fine-grained properties of the point itself. The leakage problem at the
inference interface is, at its core, that the local geometry of the decision
boundary is visible through the released label.

RP rephrases the privacy goal in exactly those geometric terms. We do not ask
that the released label be uninformative---it has to be informative, or the
service is useless---but ask instead that the label be the same for every input
inside a certified neighborhood of the queried point. Within that neighborhood,
the boundary has been pushed away from the point. The outside observer can
probe as much as they like; the released label refuses to distinguish the
queried point from any of its neighbors. In this sense, RP trades a sharp
pointwise readout for a flat plateau, and the size of the plateau is exactly
the robust radius $R$ we report as the privacy metric.

\subsection{Tuning RP: Privacy, Utility, and Inference Cost}
\label{sec:tuning-rp}

When instantiated with randomized smoothing, RP exposes two knobs to a deployer:
the noise scale $\sigma$ and the sample size $N$. The geometric picture from
Section~\ref{sec:sub:rp_intuition} explains their different roles.

The noise scale $\sigma$ primarily sets the size of the plateau the deployer is
asking for. A larger $\sigma$ typically produces a larger certified radius $R$, hence
a larger indistinguishable neighborhood around each queried input and a stronger
inference-stage privacy guarantee. The classification experiment makes this
concrete: at $N=5000$ and $\sigma=0.6$, the median compatible age interval length
left to the attacker grows from $23.50$ to $27.73$ while retaining $92.86\%$
accuracy; pushing $\sigma$ to $1.0$ increases the interval length to $29.96$,
with accuracy at $84.21\%$. The dial is monotone in privacy in this experiment, 
and it does cost utility, because asking the released label to be the same across 
a larger neighborhood asks the model to discard information that, at smaller
neighborhoods, it would have used. This part of the tradeoff is intrinsic: it
lives in the choice of how large a plateau to demand, and no amount of clever
implementation removes it.

The sample size $N$ does something different. For a fixed $\sigma$, the target
plateau is already largely specified; what $N$ controls is how faithfully the 
deployer can realize that plateau at the inference stage. A smoothed prediction 
with very few noise samples is a noisy estimate of the majority vote: near the 
boundary it can flip spuriously, certified radii can be unattainable, and 
the plateau the deployer asked for via $\sigma$ appears, in practice, as a 
noisier one. Raising $N$ tightens the estimate, lifts more queries above the
certification threshold, and lets the deployed system deliver the plateau that
$\sigma$ promised. Crucially, this happens without trading utility for privacy:
larger $N$ can improve both at once, at additional inference cost per query.
In the inversion experiment, holding $\sigma=0.08$ fixed and raising $N$ from
$10$ to $100$ takes accuracy from $78.3\%$ to $81.6\%$ and attack success rate
from $12\%$ to $8\%$ simultaneously (Figure~\ref{fig:attack-accuracy-noise}).
The classification experiment shows the same effect at a different operating
scale: at $\sigma=1.0$, raising $N$ from $1000$ to $5000$ takes test accuracy
from $81.95\%$ to $84.21\%$ and the median RAP-compatible interval length from 
$28.37$ to $29.96$ (Figure~\ref{fig:rec-rap-distribution}), again improving 
utility and privacy at once. $N$ is therefore the inference-cost budget the
deployer spends to cash in the privacy that $\sigma$ has only aimed~for.

The simple way to tune RP, then, is first to estimate $\sigma$ from a privacy
requirement: how large a plateau the application needs around each query. The
deployer can then push $N$ as far as the inference-cost budget allows.

\subsection{Handling Abstention}
\label{sec:sub:handle_abstain}

Smoothed classifiers may return an \textsc{Abstain} outcome when the top class
cannot be certified with sufficient confidence. Whether abstention should 
be exposed as part of the inference interface depends on the evaluation goal.

In the classification experiment, we measure RAP using the robust radius $R$.
We need to report abstention explicitly in this setting, because $R$ is a
certified quantity: a query has a valid radius only when the certification
confidence satisfies a user-specified lower bound $1-\alpha$. When certification
fails, there is no radius to report at this confidence level, so the query must
be counted as an abstention rather than assigned a radius it cannot certify.
Suppressing abstention would overstate the certified guarantee.

In the inversion and distillation experiments, we use an always-return-a-label
protocol. These experiments evaluate empirical attack mitigation.
Allowing abstention there would create a trivial mitigation for attacks,
especially on challenging tasks where the top-class probability under noise
often fails to reach the level required for certification and the smoothed
classifier abstains more frequently: the adversary might fail simply because the 
model refuses to answer, not because RP masks the fine-grained signals of 
input--output dependence leaked through the inference interface.

\subsection{Robust Privacy vs. Differential Privacy}
\label{sec:rp-vs-dp}

While Robust Privacy (RP) and Differential Privacy (DP)~\cite{dwork2006calibrating} 
both support privacy-preserving machine learning, they operate at different stages 
and enforce distinct notions of indistinguishability. Section~\ref{sec:instance-exp} 
gives empirical traction on what that difference looks like in practice.

DP, as typically instantiated by DP-SGD~\cite{abadi2016deep} for privacy-preserving
machine learning, asks: how much can any individual training record influence
the learned model? It enforces this bound at training time, by clipping
per-example gradients and adding noise during optimization. The released
artifact is the model itself, and the privacy guarantee is about that
artifact's constrained dependence on its training set.

RP asks a different question: how much can a single released prediction
reveal about the input that produced it? It enforces its bound at the inference
stage, by certifying that the prediction is invariant in a neighborhood of
the queried input. The released artifact is each individual prediction, and
the privacy guarantee is about that prediction's constrained dependence on 
the fine-grained details of its input.

The two notions are therefore not interchangeable, and our inversion experiments 
make the distinction concrete. The inversion attack of Kahla et al.~\cite{kahla2022label} 
obtains its signal entirely from how the released prediction shifts under 
small synthetic perturbations of the query. DP-SGD bounds memorization, 
but it does not bound the fine-grained per-query input--output dependence 
that this signal lives in: a model trained under DP-SGD can still expose an 
informative, serviceable optimization landscape to the attacker. Empirically, 
even at noise scales that drive the privacy budget $\varepsilon$ into the range 
$10^6$--$10^8$ (i.e., well past the point at which the formal DP guarantee
retains any quantitative meaning), DP-SGD trails RP in the inference-stage
privacy--utility tradeoff (Figure~\ref{fig:inversion-baseline-tradeoff}): RP 
keeps $98.4\%$ accuracy at $21\%$ attack success rate, whereas DP-SGD must 
drop accuracy to $61.7\%$ to reach a comparable attack success rate. The gap 
is not a tuning artifact, and it does not close by giving DP-SGD more privacy 
budget; it reflects the fact that the attack signal lives at the inference 
interface, not in training-stage memorization, and that an inference-stage 
defense can suppress it directly while a training-stage defense can only do 
so indirectly.

The two notions are complementary: a model can be trained under DP to bound
training-set memorization, then queried under RP to bound what each released
prediction reveals about its input. Meaningfully composing the two mechanisms
is an open direction we leave to future work.

\subsection{Scope of RP: Privacy Leakage Mitigation at Attribute and Instance Levels}
\label{sec:rp-mia-reason}

The geometric picture from Section~\ref{sec:sub:rp_intuition} also clarifies
the scope of what RP can and cannot deliver. A plateau hides what is happening
around an individual queried input, not what the underlying function does across
the whole input space. The classification and model inversion experiments fall
on the side of this distinction that RP is built to protect; the distillation
experiment sits on the other side, where the plateau abstraction no longer
applies.

Attribute inference and model inversion both succeed by exploiting fine-grained
input--output dependence of individual queries, but they use this dependence in
different ways. In attribute inference, the adversary fixes the side information
$x_{-1}$ and uses the released label to constrain the possible range of the
sensitive-attribute value; RP weakens this constraint by expanding the
RAP-compatible interval of sensitive-attribute values. In model inversion, the
adversary probes local prediction changes around a synthetic input and uses the
local boundary shape as an optimization cue for moving the synthetic input
toward the target region. RP flattens this local geometry inside the certified
neighborhood; the cue disappears, and the optimization loses its compass.

Model distillation succeeds by issuing many queries spread across the input
space and fitting a student to the global input--output behavior of the deployed
model. No single query carries the attack; the signal lives in the aggregate.
A plateau around any one query is invisible at this scale, and the global
function shape that the student needs to learn is still there to be learned.
Our experiments confirm this: distillation effectiveness tracks teacher utility, 
as expected when the attack learns the teacher's global behavior rather than 
the local per-query signal that RP suppresses. This is the opposite pattern
from attribute inference and model inversion, where attack effectiveness can
be lowered even as utility improves.

All three threats are real, and all operate through the inference interface,
but they call for defenses with different shapes. RP addresses attribute-level
and instance-level leakage from individual predictions; function-level model
extraction is better met by orthogonal mechanisms such as query budgeting,
which can be composed with RP. The lesson is not that RP is the right defense
for every inference-stage threat, but that privacy at the inference interface
needs its own abstractions and defenses. RP provides one such abstraction for
attribute-level and instance-level leakage through individual predictions.